\documentclass[copyright,creativecommons]{eptcs}

\providecommand{\event}{}
\providecommand{\volume}{??}
\providecommand{\anno}{20??}
\providecommand{\firstpage}{1}

\usepackage{breakurl}

\newtheorem{theorem}{Theorem}[section]

\title{Deterministic Autopoietic Automata}
\author{Martin F\"urer\thanks{Research supported in part by the National Science Foundation under Grant CCF-0728921}
\institute{Department of Computer Science and Engineering \\
	Pennsylvania State University \\
        University Park, PA 16802, USA}
\institute{Visiting:  Institut f\"ur Mathemtik \\
	Universit\"at Z\"urich \\
	CH-8057 ZŸrich \\
	Switzerland}
\email{furer@cse.psu.edu}
}

\def\titlerunning{Deterministic Autopoietic Automata}
\def\authorrunning{Martin F\"urer}
\begin{document}
\maketitle

\begin{abstract}
This paper studies two issues related to the paper on Computing by Self-reproduction: Autopoietic Automata by Ji{\v{r}}{\'i} Wiedermann. It is shown that all results presented there extend to deterministic computations. In particular, nondeterminism is not needed for a lineage to generate all autopoietic automata. 
\end{abstract}

\section{Introduction}
In 2001, van Leeuwen and Wiedermann \cite{vanLeeuwenW01} have defined evolving interactive systems, in particular sequences of interactive finite automata with global states, to model infinite computations on an ever changing machine or system of machines. Modern computation does not just happen on an individual machine for a fixed time, but it goes on forever over an unbounded number of software and hardware changes. Evolving automata have also been called lineage of automata \cite{VerbaanLW04}. For more background information, see \cite{Verbaan06,WiedermannL08}.

All results of this paper are related to the paper by Wiedermann \cite{Wiedermann07} studying autopoietic automata,
a special kind of offspring-producing evolving machines. The offspring relation defines trees of autopoietic automata. Attention is often focused on a lineage of autopoietic automata, a path in a tree of autopoietic automata.

Finite autopoietic automata as defined in \cite{Wiedermann07} are finite automata augmented with the following special features. They have two input options, two output options and two modes. The two modes are the reproducing mode (defined by a subset $R$ of states) and the transducer mode (defined by the complementary set $Q-R$ of the states.

In the reproducing mode the automaton uses a finite read-only input tape and a one-way output tape.
The finite automaton operates like a Turing machine. It is a finite automaton though, because the input tape is of fixed length and there are no additional work tapes.

In the transducer mode, the automaton does not change the tapes, but reads from an infinite input stream of which it can access one symbol of $\Sigma$ at a time from an input buffer, and it writes one symbol of $\Sigma$ at a time into an output buffer, producing an output stream.

During the whole operation, the input tape of the automaton $A$ actually contains the code of $A$, a straightforward description of the transition relation $\delta$. 
The code is a sequence of 5-tuples in arbitrary order. Each 5-tuple consists of an observed symbol (on the tape or in the input buffer, depending on the mode), a current state, a new symbol (to be written onto the tape or into the output buffer, depending on the mode), and a direction (to move on the input tape in reproducing mode, or a dummy indication of no move for transducer mode). For nondeteministic automata $\delta$ is an arbitrary relation, whereas we use deterministic automata here, meaning that $\delta$ is a partial function of the first two components.

Whenever the reproducing mode finishes (by going to a special state of $q_{1} \in R$, the automaton $A$ splits into 2 automata. One of them is the old $A$ with the same input tape, but with an empty output tape. The other one is the offspring $A'$, using the previous output tape as its input tape, while its new output tape is empty. 
Both automata start in the start state $q_{0} \in Q - R$ with either head at the left end on the respective tape. Depending on the application, the offspring automaton $A'$ keeps reading from the original input stream continuing at the current position (in Theorem~\ref{thm:lineage} and Theorem~\ref{thm:sim} below) or both automata receive new input streams (in Theorem~\ref{thm:all}) as in the corresponding situations in Wiedermann's paper \cite{Wiedermann07}.
At this time, the offspring automaton should have a proper encoding of a new transition function on its input tape, otherwise it stops working.

It is possible that the new automaton $A'$ is able to read from a stream containing a larger alphabet $\Sigma'$, because the encoding in binary on the input tape allows for a potentially infinite alphabet, as the symbol $\sigma_{i}$ is just encoded by $i$ (in unary representation). Naturally, as the input tape has a finite length, at any time only a constant number of symbols are allowed.




\section{The Theorems}

Interactive Turing machines as defined by \cite{vanLeeuwenW01,vanLeeuwenW01b} are Turing machines receiving a symbol at a time from a buffer connected to an input stream and writing a symbol at a time to an output buffer creating an output stream. Again, we consider deterministic interactive Turing machines. 

Recall that a lineage $\mathcal{A} = A_{1}, A_{2}, \dots $  of autopoietic automata
is a path in the tree defined by the offspring of a single autopoietic automaton $A_{1}$. When talking about a lineage of automata, we assume without loss of generality, that $A_{i+1}$ is the new offspring of $A_{i}$ rather than the replica of $A_{i}$ for every $i$. We also assume that there is just one input stream. The offspring automaton keeps reading from the position reached by the parent, even though the input stream contains more and more symbols as it reaches parts intended for later automata $A_{i}$. Thus we have just one such essential lineage for every input stream, as our automata are deterministic.

Later, we will also consider arbitrary trees obtained even in the deterministic case, by not focusing on one lineage and considering new input streams after each branching.

The following two theorems are proved exactly as in the original version \cite{Wiedermann07} with both the automaton and the Turing machine being nondeterministic. 

\begin{theorem} \label{thm:lineage}
 A lineage $\mathcal{A} = A_{1}, A_{2}, \dots $  of autopoietic automata can be simulated by an interactive Turing machine. 
\end{theorem}

The Turing machine stores the contents of the input tape of the currently simulated automaton on a tape where it can always consult it to determine the next simulation step.
Naturally, the Turing machine has fixed input and output alphabets. Therefore, it reads and writes encodings of the symbols read and written by the automata $A_{i}$.

\begin{theorem} \label{thm:sim}
Any interactive Turing machine $M$ can be simulated by a lineage of autopoietic automata.
\end{theorem}

Without loss of generality, we assume that $M$ has just one tape (infinite to the right only) and uses the alphabet $\Sigma = \{0,1,b\}$, since such a machine has the same computational power as any multi-tape machine. The automaton $A_{i}$ handles the simulation as long as the Turing machine $M$ only uses $i$ tape squares. A state of $A_{i}$ not only encodes the corresponding state of $M$, but also the tape inscription (of fixed length $i$) and the head position of $M$.

Whenever the Turing machine uses a new tape square, the simulating automaton switches to reproducing mode. It copies the part of the automaton involving the states of $R$ (handling reproduction). The part involving the states of $Q-R$ (handling transducer steps) is roughly tripled
corresponding to the additional tape square containing $0$, $1$, or $b$ (blank). Also the few additional transitions corresponding to the head being on the new square are easily handled.

More interesting is the next theorem, again corresponding to the following nondeterministic version in  \cite{Wiedermann07}. There exists an autopoietic automaton which, when working in nondeterministic input mode, generates a descendant tree containing all autopoietic automata.

Here we cannot directly extend this to our deterministic setting. We have to be careful to make sure we don't get stuck in transducer mode. We cannot produce all autopoietic automata on one path, because many such automata (on some or all input streams) never switch to reproducing mode.
Thus if we produce any such automaton, the reproduction stops on that path. 

The proof is a bit harder for deterministic automata, because the various autopoietic automata have to be created in a more systematic way. We cannot use nondeterminism to create them all. Still we have a whole tree of automata created depending on the input sequence. Thus in some sense the nondeterminism is still present in the input stream. But during the reproducing mode the input stream is not touched, allowing just one new automaton to be produced during this phase.

The essential requirement of autopoietic automata is that they have to act according to their program stored on the input tape. It is actually not clear how this is handled in proof of the nondeterministic case \cite{Wiedermann07}. In any case, nondeterminism (used during the reproducing mode) allows the creation of any possible autopoietic automaton as an offspring.

The important part of our tree of automata consists just of one infinite path
(lineage) $\mathcal{A} = A_{1}, A_{2}, \dots $  chosen as long as the input stream contains just zeros. Every automaton $A_{i}$ on this path has another child (offspring) $A'_{i}$ chosen when reading a 1. The set of automata $\{ A'_{1}, A'_{2}, \dots \}$ consists of all possible autopoietic automata. Our tree of automata does not just consist of this important part. Every $A'_{i}$ according to its definition is the root of some finite or infinite subtree.

\begin{theorem} \label{thm:all}
There exists a deterministic autopoietic automaton which depending on the input sequence generates a descendant tree containing all autopoietic automata.
\end{theorem}

\begin{proof}
We partition the states not just into two parts according to transducer mode and reproducing mode, but into four parts. The corresponding 5-tuples representing the transitions of $\delta$ are stored in 4 blocks that are easily recognized because of the different sets of states used. 

We have an active and a passive set of states for both modes. Furthermore, the active set of states for the reproducing mode is split into two pieces. Once a piece is entered, control stays in that piece until the switch to transducer mode.   As long as the input stream only consists of zeros, the lineage creates a complete variety of the two passive parts in a systematic way, with just one small deficiency. The start state has been replaced in these parts by a pseudo start state with high index. The active parts also involve only states with high indices in addition to the real start state.

As a result, as long as the input stream only provides 0, no passive state is reachable and all possible passive parts can be prepared without having to worry that the constructed automaton would have an undesired behavior. Its behavior so far is only determined by the active parts. 

We just choose a simple systematic enumeration of all automata. Repetitions are allowed. For $i=1,2,\dots$ enumerate lexicographically the (finitely many) finite automata whose states and symbols have indices at most $i$ and whose transitions are given by at most $i$ 5-tuples. Concatenate these enumerations to a single infinite enumeration of all autopoietic automata. We interpret this enumeration as an enumeration of pairs of passive parts of our automata, which in addition have basically identical active parts.
 
Now the automaton with the $j$th such pair of passive parts creates the automaton with the $(j+1)$st pair of passive parts as its offspring, after reading a single 0 from the input stream in transducer mode. The active parts of both automata agree, except for a possible increase of all the indices of states (of active parts) to allow more and more states to be used in the passive parts.

Finally, when a 1 is read from the input stream, it causes all these automata to pass control to the second piece of the active reproducing part, which has not been used so far. It causes all active parts to be erased, i.e., not being copied to the offspring. Furthermore, the pseudo start state of the parent is replaced by the real start state in the offspring. This way any possible automaton appears somewhere as an offspring.
\end{proof}

The problem of sustainable evolution asks for any autopoietic automaton and any infinite sequence of inputs whether there is an infinite lineage generated by that automaton on that input. To have a precise question, one would have to restrict attention to (in some given system)  definable  infinite sequences of inputs. But it is even undecidable for a fixed sequence.

\begin{theorem}
The problem of sustainable evolution is undecidable even for the all 0 input stream.
\end{theorem}

\begin{proof}
 By Theorem~\ref{thm:sim}, every Turing machine can be simulated by a lineage of autopoietic automata. Thus the halting problem for Turing machine can be reduced to the question whether an automaton creates an infinite lineage as follows.  A lineage is built that simulates the given Turing machine with empty input and during the simulation always creates new automata. The lineage can be defined such that it stops as soon as the simulation stops. This way the lineage is finite if and only if the Turing machine halts.
\end{proof}

In the paper \cite{Wiedermann07}, the Turing machine reduction is accidentally in the wrong direction.

\section{Conclusion}
We have shown that the autopoietic automata need not be nondeterministic to have the nice properties shown in \cite{Wiedermann07}. The construction gets only slightly more complicated.
We have pointed out the key ingredient of a proof of Theorem~\ref{thm:all} showing that some autopoietic automaton has all such automata in its subtree.

\bibliographystyle{eptcs} 

\bibliography{autopoietic}

\end{document}